\documentclass[letterpaper]{article} % DO NOT CHANGE THIS
\usepackage{aaai2026}  % DO NOT CHANGE THIS
\usepackage{times}  % DO NOT CHANGE THIS
\usepackage{helvet}  % DO NOT CHANGE THIS
\usepackage{courier}  % DO NOT CHANGE THIS
\usepackage[hyphens]{url}  % DO NOT CHANGE THIS
\usepackage{graphicx} % DO NOT CHANGE THIS
\usepackage{natbib}  % DO NOT CHANGE THIS AND DO NOT ADD ANY OPTIONS TO IT
\usepackage{caption} % DO NOT CHANGE THIS AND DO NOT ADD ANY OPTIONS TO IT
\usepackage{algorithm}
\usepackage{algpseudocode}

\usepackage{booktabs}
\usepackage{multirow}
\usepackage{bbding}
\usepackage{makecell}
\usepackage{amsmath}
\usepackage{bm}
\usepackage{amssymb}

\usepackage{newfloat}
\usepackage{listings}
\DeclareCaptionStyle{ruled}{labelfont=normalfont,labelsep=colon,strut=off} % DO NOT CHANGE THIS
\floatstyle{ruled}
\newfloat{listing}{tb}{lst}{}
\floatname{listing}{Listing}
\title{PromptMoE: Generalizable Zero-Shot Anomaly Detection \\ via Visually-Guided Prompt Mixtures}

\author {
    Yuheng Shao\textsuperscript{\rm 1,\rm 2},
    Lizhang Wang\textsuperscript{\rm 1},
    Changhao Li\textsuperscript{\rm 1},
    Peixian Chen\textsuperscript{\rm 2},
    Qinyuan Liu\textsuperscript{\rm 1, \rm 3}\thanks{Corresponding author.}
}
\affiliations {
    \textsuperscript{\rm 1}School of Computer Science and Technology, Tongji University\\
    \textsuperscript{\rm 2}Ant Group\\
    \textsuperscript{\rm 3}Key Laboratory of Embedded System and Service Computing, Ministry of Education, Tongji University\\
    
   \{yuhengshao, lizhangwang, chli, liuqy\}@tongji.edu.cn, peixian.cpx@antgroup.com
}

\usepackage{bibentry}
\begin{document}

\maketitle

\begin{abstract}
Zero-Shot Anomaly Detection (ZSAD) aims to identify and localize anomalous regions in images of unseen object classes. While recent methods based on vision-language models like CLIP show promise, their performance is constrained by existing prompt engineering strategies. Current approaches, whether relying on single fixed, learnable, or dense dynamic prompts, suffer from a representational bottleneck and are prone to overfitting on auxiliary data, failing to generalize to the complexity and diversity of unseen anomalies. To overcome these limitations, we propose $\mathtt{PromptMoE}$. Our core insight is that robust ZSAD requires a compositional approach to prompt learning. Instead of learning monolithic prompts, $\mathtt{PromptMoE}$ learns a pool of expert prompts, which serve as a basis set of composable semantic primitives, and a visually-guided Mixture-of-Experts (MoE) mechanism to dynamically combine them for each instance.
Our framework materializes this concept through a Visually-Guided Mixture of Prompt (VGMoP) that employs an image-gated sparse MoE to aggregate diverse normal and abnormal expert state prompts, generating semantically rich textual representations with strong generalization. Extensive experiments across 15 datasets in industrial and medical domains demonstrate the effectiveness and state-of-the-art performance of $\mathtt{PromptMoE}$.
\end{abstract}


\section{Introduction}
Anomaly detection is a crucial technology with widespread applications, such as ensuring quality in industrial manufacturing~\cite{bergmann2019mvtec, jezek2021deep, zou2022spot, wang2024real} and assisting in medical diagnosis~\cite{buda2019association, huang2024adapting}. For example, in automated production lines, models must accurately detect subtle defects like scratches, blemishes, or deformations. In practice, however, new product categories and unforeseen defect types frequently emerge, making it infeasible to collect extensive labeled data or retrain models for every new class. This has highlighted the importance of Zero-Shot Anomaly Detection (ZSAD), which aims to detect anomalies in object categories that are not seen during training. ZSAD is both essential and inherently challenging.

\begin{figure}[t]
        % \centering
	\includegraphics[width=.95\linewidth]{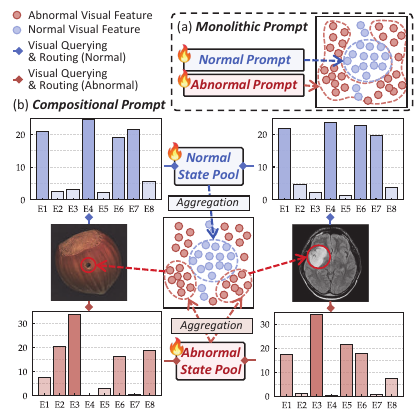}
	\caption{(a) Monolithic Prompt vs. (b) Compositional Prompt. Unlike monolithic prompts with fixed representations, our compositional prompts employs visual querying to dynamically aggregate expert prompts, treated as semantic primitives, into an instance-specific textual representation.}
\label{fig:teaser}
\end{figure}

Recently, the emergence of large-scale vision-language models, particularly CLIP~\cite{radford2021learning}, offers new possibilities for ZSAD due to their strong generalization and zero-shot recognition capabilities. CLIP, trained on massive image-text pairs, learns a shared semantic space that aligns visual and textual information. Motivated by this, recent studies have explored adapting CLIP for ZSAD. Some methods~\cite{winclip, chen2023april} reformulate ZSAD as an image-text matching task using hand-crafted prompts to describe ``normal'' and ``abnormal'' states. To reduce the need for manual design, later methods like AnomalyCLIP~\cite{anomalyclip} introduced learnable prompts that are optimized on auxiliary datasets.

Although prompt-based methods have shown promising results, they face notable limitations in representational power and generalization.
Firstly, the single-prompt strategy, whether manually designed or learned as a single normal/abnormal pair, suffers from a clear representational bottleneck, as a fixed prompt vector struggles to capture the diverse patterns of normality and abnormality in unseen classes.
Secondly, simply increasing the number of learnable static prompts is not a viable solution, as it significantly increases the risk of overfitting to the auxiliary data; the model tends to memorize specific pattern-prompt combinations from the training set instead of learning generalizable abstract concepts.
Furthermore, approaches~\cite{adaclip, vcpclip} inspired by CoCoOp~\cite{zhou2022conditional} that dynamically generate prompts from visual instances often rely on a single mapping network to handle all visual variations, making it difficult to generate specialized prompts for specific, fine-grained anomaly patterns.
Finally, the overfitting issue inherent in static designs persists even in deep prompting approaches. As demonstrated by previous work~\cite{anomalyclip}, when static learnable prompts are inserted into the text encoder's intermediate layers, they are effective only in shallow stages. Applying them to deeper layers harms zero-shot detection on unseen classes, revealing a critical generalization failure of static prompt designs.

To address the aforementioned challenges, we propose $\mathtt{PromptMoE}$, a novel framework that shifts prompt learning from a monolithic to a compositional paradigm, as conceptually illustrated in Fig.~\ref{fig:teaser}.
Our key insight is that to effectively represent the open-ended and diverse nature of normal and abnormal patterns in ZSAD, a compositional representation capability for prompts is key, rather than learning a fixed or single dynamic prompt.
Accordingly, $\mathtt{PromptMoE}$ shifts the paradigm from learning a monolithic state description to learning a basis set of composable semantic primitives (i.e., expert prompts) and a router to dynamically combine them based on the visual instance. This Mixture-of-Experts (MoE) approach allows for the creation of a rich variety of instance-specific prompts from a finite set of parameters, while the sparse activation inherent to MoE helps mitigate overfitting. 
Our framework implements this strategy through a novel \textbf{Visually-Guided Mixture of Prompt (VGMoP)} module, which employs an image-gated sparse MoE for specialized selection from distinct expert pools to construct fine-grained, instance-adaptive textual prompts. This process is further regularized by expert load balancing and decoupling losses to enhance the diversity of the expert pool, thereby improving generalization to unseen classes.

Our main contributions are summarized as follows:
\begin{itemize}
    \item We propose $\mathtt{PromptMoE}$, a framework for ZSAD that replaces monolithic prompts with a compositional learning paradigm. It employs a MoE mechanism to dynamically combine a learned basis of semantic primitives, enhancing generalization and mitigating overfitting.
    \item We design a VGMoP module to realize this paradigm. It utilizes an image-gated sparse router and distinct expert pools to dynamically compose specialized normal and abnormal state prompts for each visual instance.

    \item Extensive experiments 15 industrial and medical datasets demonstrate that our method achieves SOTA performance, validating its effectiveness on the ZSAD task.
\end{itemize}

\begin{figure}[tbhp]
        \centering
	\includegraphics[width=.95\linewidth]{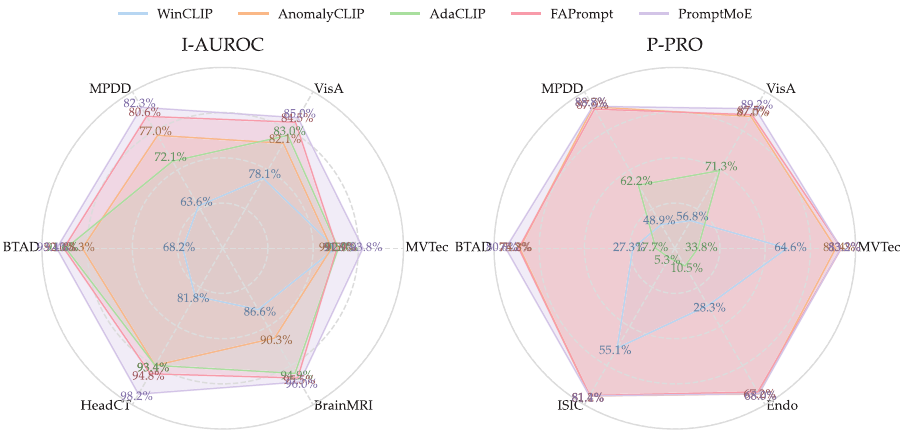}
	\caption{ZSAD performance of $\mathtt{PromptMoE}$ compared to state-of-the-art methods. Left: I-AUROC. Right: P-PRO.}
	\label{fig:radar}
\end{figure}

\section{Related Work}
\subsection{Zero-Shot Anomaly Detection}

Recent research in ZSAD has been significantly advanced by Vision-Language Models (VLM) like CLIP~\cite{radford2021learning}, which typically reframe the task as image-text matching.
Early pioneering works, such as WinCLIP~\cite{winclip}, introduce a prompt ensembling strategy, generating rich textual descriptions by combining numerous hand-crafted templates with state words. To overcome the reliance on manual design, AnomalyCLIP~\cite{anomalyclip} proposes prompt learning, which optimizes learnable vectors to replace fixed textual prompts, thereby automatically learning general anomaly concepts.

However, static learnable prompts show limited adaptability to different visual instances, leading subsequent research toward dynamic, visually-guided prompt generation. For instance, AdaCLIP~\cite{adaclip} combines static and dynamic prompts, while VCP-CLIP~\cite{vcpclip} injects visual context at multiple stages of the text encoder. Another approach, taken by Bayes-PFL~\cite{qu2025bayesian}, models the prompt space as a learnable probability distribution and generates diverse prompts via sampling. Similarly, FAPrompt~\cite{zhu2024fine} learns a set of fine-grained prompts and adapts them using a visual prior extracted from the test image. While these methods enhance instance-adaptability, they often rely on complex generative or probabilistic modeling. In contrast, our work explores using a MoE mechanism to achieve visually-guided prompt generation in a compositional manner.

\begin{figure*}[t]
	\centering
	\includegraphics[width=0.8\textwidth]{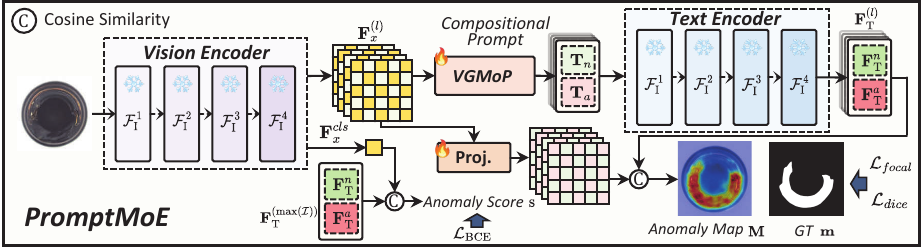}
	\caption{Framework of $\mathtt{PromptMoE}$. }
	\label{fig:framework}
\end{figure*}

\subsection{Mixture of Experts}
Mixture-of-Experts is a classic conditional computation paradigm that effectively scales model capacity without a proportional increase in computational cost by activating only a sparse subset of ``expert'' subnetworks for each input~\cite{MoE}. This strategy has recently been widely and successfully applied to scale large-scale language models to trillions of parameters~\cite{switchtrans} and has also been adapted for vision models~\cite{riquelme2021scaling}. In these mainstream applications, MoE typically functions as a dynamic router that selects one of several expert MLP layers to process each input token. In contrast, our $\mathtt{PromptMoE}$ framework explores a novel application of MoE in prompt engineering: we re-purpose the MoE mechanism as a strategy for compositional prompt generation. Under this framework, the ``experts'' are learnable prompt embedding segments, and an image-gated router learns to select a specialized subset of these experts to dynamically construct a highly adaptive textual representation for each visual instance.

\section{Method}

\subsection{Problem Formulation} 

The goal of ZSAD is to detect and segment anomalous regions in images, especially for object classes that are unseen during training. Formally, given an input image $\mathbf{x} \in \mathbb{R}^{h \times w \times 3}$, the model aims to predict an image-level anomaly score $s$ as well as a pixel-level anomaly map $\mathbf{M} \in \mathbb{R}^{h \times w}$.
We utilize an auxiliary training dataset $\mathcal{D}_{train} = \{(\mathbf{x}_i, c_i, \mathbf{m}_i)\}_{i=1}^{N_{train}}$, where each sample consists of an image $\mathbf{x}_i$, an image-level label $c_i$, and a pixel-level anomaly mask $\mathbf{m}_i$. The label $c_i \in \{0, 1\}$ indicates whether the image is normal ($c_i=0$) or abnormal ($c_i=1$),  and the mask $\mathbf{m}_i \in \{0,1\}^{h \times w}$ provides pixel-wise annotations for anomalous regions. The test set $\mathcal{D}_{test} = \{\mathbf{x}_j\}_{j=1}^{N_{test}}$ contains images from unseen classes that are disjoint from those in the training set, i.e., $\mathcal{C}_{train} \cap \mathcal{C}_{test} = \emptyset$. The model is trained to learn a generalizable patterns of normality and abnormality from $\mathcal{D}_{train}$, and to achieve the detection of anomalies in unseen classes within $\mathcal{D}_{test}$.

% \subsection{Motivation}

% To achieve this, we address two key challenges:
% \begin{itemize}
%     \item we

%     \item we
% \end{itemize}
\begin{figure}[tbhp]
	\includegraphics[width=1\linewidth]{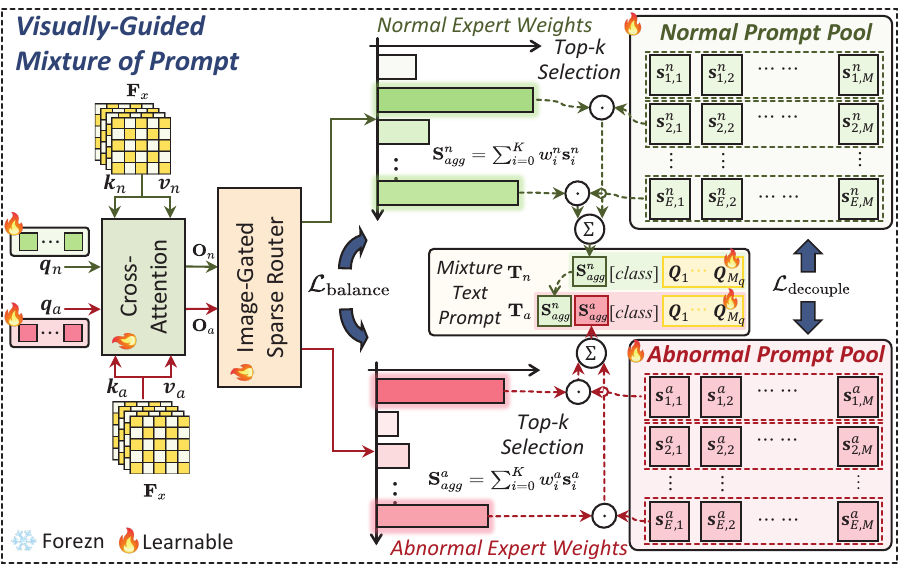}
	\caption{Architecture of our proposed Visually-Guided Mixture of Prompt (VGMoP) module. A sparse router, guided by cross-attention between learnable queries and image features, dynamically selects and aggregates top-k experts from respective prompt pools to construct instance-specific normal and abnormal text prompts.}
	\label{fig:vgmop}
\end{figure}

\subsection{Overview} 
We propose $\mathtt{PromptMoE}$, a framework that utilizes visually-guided prompt mixtures to improve generalization to unseen classes, thereby enhancing ZSAD performance. 
As illustrated in Fig.~\ref{fig:framework}, given an input image $\mathbf{x}$, a vision encoder extracts  comprehensive patch-level features $\mathbf{F}_x^{(l)} \in \mathbb{R}^{(HW+1) \times D_{x}}$ and a global feature $\mathbf{F}_x^{cls} \in \mathbb{R}^{D_{x}}$, where $H$ and $W$ denote the number of patches, and $D_x$ is the feature dimension.
The core innovation of $\mathtt{PromptMoE}$ is the VGMoP module, detailed in Fig.~\ref{fig:vgmop}. It takes the visual features $\mathbf{F}_x^{(l)}$ as input and dynamically generates fine-grained, instance-specific normal and abnormal textual prompts, $\mathbf{T}_n^{(l)}$ and $\mathbf{T}_a^{(l)}$, for each visual instance.
Finally, layer-wise similarities between textual embeddings $\mathbf{F}_\text{T}^{(l)}$ and patch features $\mathbf{F}_x^{(l)}$ are aggregated to produce the anomaly map $\mathbf{M}$.

\subsection{Visually-Guided Mixture of Prompt}
Previous ZSAD method \cite{anomalyclip} introduces learnable pairs of normal and abnormal text prompts to guide pre-trained CLIP models in distinguishing these two concepts, thereby avoiding tedious manual template design. However, such single, fixed prompts possess limited expressive power, struggling to adequately represent the complex and diverse anomaly patterns found in unseen classes. Although some subsequent studies \cite{adaclip, vcpclip} attempt to dynamically generate prompts using visual features to enhance instance adaptability, their effectiveness is often constrained by the inherent representational bottlenecks of visual encoders. Furthermore, simply increasing the number of prompts or their complexity can easily lead to model overfitting on seen classes, consequently impairing its generalization ability to unseen classes.

To address these limitations, we propose VGMoP, a visually-guided prompt mixture that effectively utilizes richer prompt information to enhance generalization to unseen classes.
Unlike fixed or single dynamically-generated prompts, VGMoP employs an image-gated sparse MoE mechanism. This adaptively selects and aggregates a sparse subset of relevant expert prompts from a pool for the current visual query instance, thereby constructing distinct textual representations for normal and abnormal states.

\subsubsection{Mixture Text Prompt Structure.}
Under the VGMoP strategy, we construct two mixture text prompts for each visual instance $\mathbf{x}$: a normal-state prompt $\mathbf{T}_{n} \in \mathbb{R}^{N_n \times D}$ and an abnormal-state prompt $\mathbf{T}_{a} \in \mathbb{R}^{N_a \times D}$, where $D$ denotes the token embedding dimension. The prompt are defined as:

\begin{equation}
\begin{aligned}
\mathbf{T}_{n} &= [\mathbf{S}_\text{agg}^n] \mathtt{[cls]}[\mathbf{Q}_\text{ctx}],\\ 
\mathbf{T}_{a} &= [\mathbf{S}_\text{agg}^n][\mathbf{S}_\text{agg}^a] \mathtt{[cls]}[\mathbf{Q}_\text{ctx}],
\end{aligned}
\label{prompt}
\end{equation}
where $\mathbf{S}_\text{agg}^n \in \mathbb{R}^{M_n \times D}$ represents the aggregated normal state. Inspired by \citet{promptad}, we append an aggregated abnormal state suffix $\mathbf{S}_\text{agg}^a\in \mathbb{R}^{M_a \times D}$ to $\mathbf{S}_\text{agg}^n$ to construct the abnormal-state prompt. Both $\mathbf{S}_\text{agg}^n$ and $\mathbf{S}_\text{agg}^a$ are generated by the visually-guided MoE mechanism. $\mathbf{Q}_\text{ctx} \in \mathbb{R}^{M_q \times D}$ denotes a set of learnable context tokens shared by both prompts, and $\mathtt{[cls]}$ denotes the embedding of a given category name or a generic placeholder such as ``object". The resulting $\mathbf{T}_{n}$ and $\mathbf{T}_{a}$ provide the text encoder with dynamically adapted, instance-specific semantic guidance.

\subsubsection{Visually-Guided MoE for State Aggregation.}
To achieve effective compositional representations and enhance generalization to unseen classes, $\mathbf{S}^n_\text{agg}$ and $\mathbf{S}^a_\text{agg}$ are generated using a visually-guided MoE procedure applied independently for each layer $l \in\mathcal{I}$. For each layer, the process begins by constructing a comprehensive visual representation $\mathbf{F}_x^{(l)} \in \mathbb{R}^{(HW+1) \times D_x}$, obtained by concatenating the patch features of the layer with the global feature $\mathbf{F}_x^{cls}$.
A set of learnable, layer-specific state queries $\mathbf{q}^{(l)} \in \mathbb{R}^{N_q \times D}$ then attend to $\mathbf{F}_x^{(l)}$ via a cross-attention layer to extract fine-grained, state-relevant visual contextual representations $\mathbf{O}^{(l)} \in \mathbb{R}^{N_q \times D}$, computed as:

\begin{equation}
\begin{aligned}
Q^{(l)} &= \mathbf{q}^{(l)}, K^{(l)}=\mathbf{F}_x^{(l)} \mathbf{W}_K^{(l)}, V^{(l)}= \mathbf{F}_x^{(l)} \mathbf{W}_V^{(l)},\\
\mathbf{O}^{(l)} &= \text{Softmax}\left(\frac{Q^{(l)} {K^{(l)}}^\top}{\sqrt{D}}\right) V^{(l)},
\label{cross-attention}
\end{aligned}
\end{equation}
where $\mathbf{W}_K^{(l)}, \mathbf{W}_V^{(l)} \in \mathbb{R}^{D_x \times D}$ are linear projection matrices mapping $\mathbf{F}^{(l)}_x$ to dimension $D$. 
The routing representation is then obtained by average pooling: $\mathbf{r}^{(l)} = \text{mean}(\mathbf{O}^{(l)})\in \mathbb{R}^D$.

This representation $\mathbf{r}^{(l)}$ is fed into a layer-specific image-gated sparse router $G^{(l)}(\cdot)$, composed of a two-layer MLP (i.e., $\text{Linear-ReLU-Linear}$). The router produces routing logits $\mathbf{z}^{(l)}$ for the corresponding layer-specific expert prompt pool $\mathcal{E}^{(l)} = \{\mathbf{s}^{(l)}_j \in \mathbb{R}^{M \times D}\}_{j=1}^{E}$.
From these logits, we select the top-$k$ values, denoted as $\mathbf{z}^{(l)}_{\text{top}}$, and retrieve their associated expert prompts $\{\mathbf{s}^{(l)}_{\text{top},i}\}_{i=1}^{k}$. Normalized gating weights are then computed as $\mathbf{w}^{(l)} = \text{Softmax}(\mathbf{z}^{(l)}_{\text{top}})$, which are used to aggregate the layer-specific state prompt $\mathbf{S}_\text{agg}^{(l)}$:

\begin{equation}
\mathbf{S}_\text{agg}^{(l)} = \sum_{i=1}^{k} \mathbf{w}^{(l)}_{i}\mathbf{s}^{(l)}_{\text{top},i}.
\label{eq:expert_aggregation}
\end{equation}

Note that we utilize independent expert prompt pools for normal and abnormal states, $\mathcal{E}_n$ and $\mathcal{E}_a$ respectively, and apply this Visually-Guided MoE procedure to each.
This separated design is intended to provide specialized representational spaces for normality and abnormality, thereby enhancing generalization to complex patterns in unseen classes.

\subsubsection{Auxiliary Loss.}
For each visual instance, VGMoP selects $k$ expert state prompts from the expert pool. However, during training, prompts that perform better in the early stages are more likely to be repeatedly selected by the sparse router, leading to imbalanced prompt utilization. This imbalance may cause the prompt pool to degenerate into a fixed combination of prompt states, undermining the dynamics and sparsity of selection.
Drawing inspiration from prior research ~\cite{MoE,switchtrans}, this loss encourages a uniform average contribution from all experts across the batch, defined as:

\begin{equation}
\begin{aligned}
\mathcal{L}_\text{balance} &= \alpha \sum_{l \in \mathcal{I}} \left( E \sum_{j=1}^E \left(\frac{1}{B}\sum_{i=1}^{B} \mathbf{p}^{(l)}_{i,j}\right)^2 \right), \\
\mathbf{p}^{(l)} &=\text{Softmax}(\mathbf{z}^{(l)}),
\end{aligned}
\label{eq:balance_loss}
\end{equation}
where $B$ is the batch size, $E$ is the number of experts, $\mathbf{p}^{(l)}$ is the router's output probability distribution over experts at layer $l$, and $\alpha$ is the loss weight. In practice, we find this loss is most effective when combined with a learning rate warmup schedule during the initial training epochs.

To further enhance the generalization ability of state prompts, we introduce an expert decoupling loss, $\mathcal{L}_\text{decouple}$, aimming to promote representational diversity within the expert prompt pool, ensuring that individual state prompt learn discriminative features.
Let $\bar{\mathbf{s}}_j^{(l)} = \text{mean}(\mathbf{s}_j^{(l)}) \in \mathbb{R}^D$ denote the mean embedding representation of each state prompt $\mathbf{s}_j^{(l)}$ from the prompt pool $\mathcal{E}^{(l)}$. These are then normalized to yield $\hat{\mathbf{s}}_j = \bar{\mathbf{s}}_j / \|\bar{\mathbf{s}}_j\|_2$. The normalized mean representations $\hat{\mathbf{s}}_j^{(l)}$ form the rows of a matrix $\hat{\mathbf{S}}^{(l)} \in \mathbb{R}^{E \times D}$. The expert decoupling loss $\mathcal{L}_\text{decouple}$ is defined as:

\begin{equation}
\mathcal{L}_\text{decouple} = \beta \sum_{l \in \mathcal{I}} \| \hat{\mathbf{S}}^{(l)} (\hat{\mathbf{S}}^{(l)})^T - \mathbf{I}_E \|_F^2,
\label{eq:decouple_loss}
\end{equation}
where $\beta$ is a hyperparameter weight, $\mathbf{I}_E$ is the identity matrix, and $\|\cdot\|_F^2$ denotes the squared Frobenius norm. This loss promotes orthogonality among the mean representations of different expert prompts within each layer's pool, enhancing the overall model's diversity.

\subsection{Discussion}
\label{sec:discussion}
In summary, the design of our $\mathtt{PromptMoE}$ framework addresses several key considerations. First, to effectively utilize the rich yet redundant patch features from the visual encoder, we employ a query-driven cross-attention mechanism. This mechanism learns to actively distill state-relevant visual signals from the numerous patches, avoiding the loss of critical local information that can occur with simple average pooling. Second, this query-and-distill mechanism is extended layer-wise, enabling the generation of fine-grained textual prompts that correspond to visual features at diverse semantic depths. Finally, dedicated auxiliary losses are introduced to ensure the MoE system trains stably and generalizes effectively: a load balancing loss maintains the router's dynamism, while a decoupling loss enhances generalization to unseen classes by promoting expert diversity.

\subsection{Training and Inference}
$\mathtt{PromptMoE}$ computes the pixel-level anomaly map $\mathbf{M}$ by aggregating similarity maps over a predefined set of visual encoder layers $\mathcal{I}$.  
For each layer $l \in \mathcal{I}$, the map is derived from the similarity between its patch features $\mathbf{F}_x^{(l)}$ and the corresponding textual embedding $\mathbf{F}_\text{T}^{(l)}$:

% PromptMoE computes a pixel-level anomaly map $\mathbf{M}$ by aggregating layer-wise similarities between patch features $\mathbf{F}_x^{(l)}$ and textual embeddings $\mathbf{F}_\text{T}$:
\begin{equation}
\mathbf{M} = \sum_{l \in \mathcal{I}}\text{Softmax}(\text{Up}(\mathbf{F}_x^{(l)}{\mathbf{F}_\text{T}^{(l)}}^\top)/\tau),
\label{eq:map}
\end{equation}
where $\text{Up}(\cdot)$ upsamples the similarity map to $(h, w)$ and $\tau$ is a temperature parameter.
The image-level anomaly score $s$ combines the peak value of the anomaly map $\mathbf{M}$ with the similarity between the global image feature $\mathbf{F}_x^{cls}$ and the final layer's textual embedding $\mathbf{F}_\text{T}^{(\max(\mathcal{I}))}$:
\begin{equation}
s = \frac{1}{2}(\text{max}(\mathbf{M}) + \text{Softmax}(\mathbf{F}_x^{cls}{\mathbf{F}_\text{T}^{(\max(\mathcal{I}))}}^\top / \tau')).
\label{eq:imagescore}
\end{equation}

Similar to previous works~\cite{anomalyclip, gu2024anomalygpt, adaclip}, $\mathtt{PromptMoE}$ supervises the anomaly map $\mathbf{M}$ using Dice~\cite{diceloss} and Focal~\cite{focalloss} losses against the ground truth mask $\mathbf{m}$. In addition, the anomaly score $s$ is optimized using BCE loss with the ground truth label $c$. The overall loss $\mathcal{L}_\text{total}$ is:
\begin{equation}
\begin{aligned}
\mathcal{L}_\text{total} = \ 
& \underbrace{\text{BCE}(s, c)}_{\text{Classify}} + \underbrace{\mathcal{L}_\text{balance} + \mathcal{L}_\text{decouple}}_{\text{Auxiliary}}\\
+\, & \underbrace{\text{Dice}(\mathbf{M}, \mathbf{m}) + \text{Focal}(\mathbf{M}, \mathbf{m})}_{\text{Segment}}.\\
\end{aligned}
\label{eq:loss}
\end{equation}

\begin{table*}[thb]   
	\centering
        \small
	{
		\setlength\tabcolsep{1mm} 
		\begin{tabular}{cccccccccc}
			\toprule
			& Metric & Datasets & CLIP & WinCLIP & APRIL-GAN & AnomalyCLIP & AdaCLIP & FAPrompt & $\mathtt{PromptMoE}$ \\
			\midrule
			\multirow{16}{*}{\rotatebox[origin=c]{90}{Industrial}} 
			& \multirow{8}{*}{\makecell[c]{Image-level \\ (AUROC,  AP)}} 
			& MVTec AD & (74.1, 87.6) & (91.8, \underline{96.5}) & (86.1, 93.5) & (91.5, 96.2) & (\underline{92.0}, 96.4) & (91.9, 95.7) & (\textbf{93.8}, \textbf{97.2})  \\ % 8-4-8-8-5-6-diff-multilayer-nostatic-visa
			&  & VisA & (66.4, 71.5) & (78.1, 81.2) & (78.0, 81.4) & (82.1, 85.4) & (83.0, 84.9) & (\underline{84.5}, \underline{86.8}) & (\textbf{85.0}, \textbf{88.2}) \\ % 8-4-8-diff-multilayer-mvtec
			&  & MPDD & (54.3, 65.4) & (63.6, 69.9) & (73.0, 80.2) & (77.0, 82.0) & (72.1, 77.6) & (\underline{80.6}, \underline{83.3}) & (\textbf{82.3}, \textbf{84.5}) \\
			&  & BTAD & (34.5, 52.5) & (68.2, 70.9) & (73.6, 68.6) & (88.3, 87.3) & (91.6, \underline{92.4}) & (\underline{92.0}, 92.2) & (\textbf{93.4}, \textbf{95.7}) \\
			&  & SDD  & (65.7, 45.2) & (84.3, 77.4) & (79.8, 71.4) & (84.7, 80.0) & (81.2, 72.6) & (\textbf{98.6}, \textbf{95.9})  &(\underline{97.4}, \underline{94.0}) \\
			&  & DAGM & (79.6, 59.0) & (91.8, 79.5) & (94.4, 83.8) & (97.5, 92.3) & (96.5, \underline{95.7})& (\underline{98.9}, \underline{95.7}) & (\textbf{98.9}, \textbf{96.4}) \\
			&  & DTD-Synthetic & (71.6, 85.7) & (93.2, 92.6) & (86.4, 95.0) & (\underline{93.5}, 97.0) & (92.8, 97.0) &  (\textbf{95.9}, \textbf{98.3}) & (\textbf{95.9}, \underline{98.1}) \\
            \cmidrule{3-10}
            &  & Average & (63.7, 66.7) & (81.6, 81.1) & (81.6, 82.0) & (87.8, 88.6) & (87.0, 88.0) &  (\underline{91.7}, \underline{92.5}) & (\textbf{92.4}, \textbf{93.4}) \\
			\cmidrule{2-10}
			& \multirow{8}{*}{\makecell[c]{Pixel-level \\ (AUROC, PRO)}}   
			& MVTec AD & (38.4, 11.3) & (85.1, 64.6) & (87.6, 44.0) & (91.1, 81.4) & (86.8, 33.8) &  (\underline{90.6}, \textbf{83.3}) & (\textbf{91.8}, \underline{83.2}) \\
			&  & VisA & (46.6, 14.8) & (79.6, 56.8) & (94.2, 86.8) & (95.5, 87.0) & (95.1, 71.3) &  (\textbf{95.9}, \underline{87.5}) & (\underline{95.6}, \textbf{89.2}) \\
			&  & MPDD & (62.1, 33.0) & (76.4, 48.9) & (94.1, 83.2) & (\underline{96.5}, \underline{88.7}) & (96.4, 62.2) & (\underline{96.5}, 87.9) & (\textbf{96.8}, \textbf{88.8}) \\
			&  & BTAD & (30.6, 4.4) & (72.7, 27.3) & (60.8, 25.0) & (94.2, 74.8) & (87.7, 17.7) &  (\textbf{95.6}, \underline{75.2}) & (\underline{94.9}, \textbf{80.6}) \\
			&  & SDD & (39.0, 8.9) & (68.8, 24.2) & (79.8, 65.1) & (90.6, 67.8) & (71.7, 17.6) & (\textbf{98.3}, \underline{93.6}) & (\underline{98.1}, \textbf{95.6}) \\
			&  & DAGM & (28.2, 2.9) & (87.6, 65.7) & (82.4, 66.2) & (95.6, 91.0) & (97.0, 40.9) &  (\textbf{98.3}, \textbf{95.4}) & (\underline{97.8}, \underline{93.9}) \\
			&  & DTD-Synthetic & (33.9, 12.5) & (83.9, 57.8) & (95.3, 86.9) & (\underline{97.9}, 92.3) & (94.1, 24.9) &  (\textbf{98.3}, \underline{93.1}) & (\textbf{98.3}, \textbf{93.2}) \\
            \cmidrule{3-10}
            &  & Average & (39.8, 12.5) & (79.1, 49.3) & (84.8, 65.3) & (94.5, 83.3) & (89.8, 38.4) &  (\textbf{96.2}, \underline{88.0}) & (\textbf{96.2}, \textbf{89.2}) \\
			\midrule
			\multirow{10}{*}{\rotatebox[origin=c]{90}{Medical}}
			& \multirow{4}{*}{\makecell[c]{Image-level \\ (AUROC, AP)}} 
			& HeadCT & (56.5, 58.4) & (81.8, 80.2) & (89.1, 89.4) & (93.4, 91.6) & (93.4, 92.2) &  (\underline{94.8}, \underline{93.5}) & (\textbf{98.2}, \textbf{98.2}) \\
			&  & BrainMRI & (73.9, 81.7) & (86.6, 91.5) & (89.3, 90.9) & (90.3, 92.2) & (94.9, 94.2) &  (\underline{95.5}, \underline{95.6}) & (\textbf{96.0}, \textbf{96.6}) \\
			&  & Br35H & (78.4, 78.8) & (80.5, 82.2) & (93.1, 92.9) & (94.6, 94.7) & (95.7, 95.7) &  (\underline{97.8}, \underline{97.5}) & (\textbf{97.9}, \textbf{97.7}) \\
            \cmidrule{3-10}
            &  & Average & (69.6, 73.0) & (83.0, 84.6) & (90.5, 91.1) & (92.8, 92.8) & (94.7, 94.0) &  (\underline{96.0}, \underline{95.5}) & (\textbf{97.4}, \textbf{97.5}) \\
		\cmidrule{2-10}
			& \multirow{6}{*}{\makecell[c]{Pixel-level \\ (AUROC, PRO)}} 
			& ISIC & (33.1, 5.8) & (83.3, 55.1) & (89.4, 77.2) & (89.7, 78.4) & (85.4, 5.3) &  (\underline{90.9}, \underline{81.2}) &(\textbf{91.1}, \textbf{81.4}) \\
			&  & CVC-ColonDB & (49.5, 15.8) & (70.3, 32.5) & (78.4, 64.6) & (81.9, 71.3) & (79.3, 6.5) &  (\textbf{84.6}, \underline{74.7}) & (\underline{84.3}, \textbf{74.9}) \\
			&  & CVC-ClinicDB & (47.5, 18.9) & (51.2, 13.8) & (80.5, 60.7) & (82.9, 67.8) & (\textbf{85.3}, 14.6) &  (\underline{84.7}, \underline{70.1}) & (84.5, \textbf{70.9}) \\
			&  & Kvasir & (44.6, 17.7) & (69.7, 24.5) & (75.0, 36.2) & (78.9, 45.6) & (79.4, 12.3)  &  (81.2, 47.8) & \textbf{(81.6}, \textbf{48.8}) \\
			&  & Endo & (45.2, 15.9) & (68.2, 28.3) & (81.9, 54.9) & (84.1, 63.6) & (84.0, 10.5) &  (\textbf{86.4}, 67.2) & (\underline{86.3}, \textbf{68.0}) \\
            \cmidrule{3-10}
            &  & Average & (44.0, 14.8) & (68.5, 30.8) & (81.0, 58.2) & (\underline{83.5}, 65.3) & (82.7, 9.8) &  (\textbf{85.6}, \underline{68.2}) & (\textbf{85.6}, \textbf{68.8}) \\
			\bottomrule
		\end{tabular}
	}
\caption{ZSAD performance comparison on industrial and medical domains. Best results are in \textbf{bold}, second-best are \underline{underlined}.}
\label{table:zsad_all}

\end{table*}

\section{Experiments}
\subsection{Experiment Settings}
\subsubsection{Datasets.} 
To comprehensively evaluate the generalization capability of our proposed $\mathtt{PromptMoE}$ on the ZSAD task, we conduct extensive experiments on 15 real-world datasets spanning both industrial manufacturing and medical diagnosis domains. We utilize seven widely-used industrial benchmarks, including MVTec AD~\cite{bergmann2019mvtec}, VisA~\cite{zou2022spot}, MPDD~\cite{jezek2021deep}, BTAD~\cite{mishra2021vt}, SDD~\cite{tabernik2020segmentation}, DAGM~\cite{wieler2007weakly}, and DTD-Synthetic~\cite{aota2023zero}; as well as eight medical datasets from diverse modalities and application scenarios, including HeadCT~\cite{salehi2021multiresolution}, BrainMRI~\cite{kanade2015brain}, Br35H~\cite{hamada2020br35h}, ISIC~\cite{codella2018skin}, CVC-ColonDB~\cite{tajbakhsh2015automated}, CVC-ClinicDB~\cite{bernal2015wm}, Kvasir~\cite{jha2019kvasir}, and Endo~\cite{hicks2021endotect}. Following the standard ZSAD setting, we primarily train our model on the MVTec AD dataset and then perform zero-shot inference and evaluation on the other 14 datasets. To evaluate performance on MVTec AD itself, we train our model on VisA, a dataset with a disjoint set of object categories.

\subsubsection{Metrics.}
To evaluate the performance of our model, we follow the standard protocols used in previous works~\cite{anomalyclip}. For the image-level anomaly detection, we report the AUROC and AP. For the pixel-level anomaly localization, we use pixel-wise AUROC and the area under the Per-Region Overlap curve (PRO)~\cite{bergmann2019mvtec}.

\subsubsection{Implementation details.}
We utilize the pre-trained CLIP \cite{radford2021learning} (ViT-L/14@336px) released by OpenAI as our backbone, with all of its parameters frozen during training. All input images are resized to $518 \times 518$ pixels. We extract patch features from the \{6, 12, 18, 24\}-th layers of the 24-layer visual encoder. The model is trained for 15 epochs using the Adam optimizer with a batch size of 16 and a learning rate of 0.001, employing a warm-up strategy during the first 3 epochs.

For our VGMoP module, we set the number of queries $N_q=8$ and experts $E=8$, from which we sparsely select the top $k=4$ for aggregation; its cross-attention utilizes 8 heads and its router has a hidden dimension of 256. The resulting normal and abnormal state sequences have lengths $M_n=5$ and $M_a=6$, respectively, while the shared context $\mathbf{Q}_{ctx}$ has a length of $M_q=8$. The auxiliary loss coefficients $\alpha$ and $\beta$ are set to 0.01 and 0.005, respectively. All experiments are conducted on a RTX3090 using PyTorch 1.13.1.

\begin{figure}[bth]
        \centering
	\includegraphics[width=1.\linewidth]{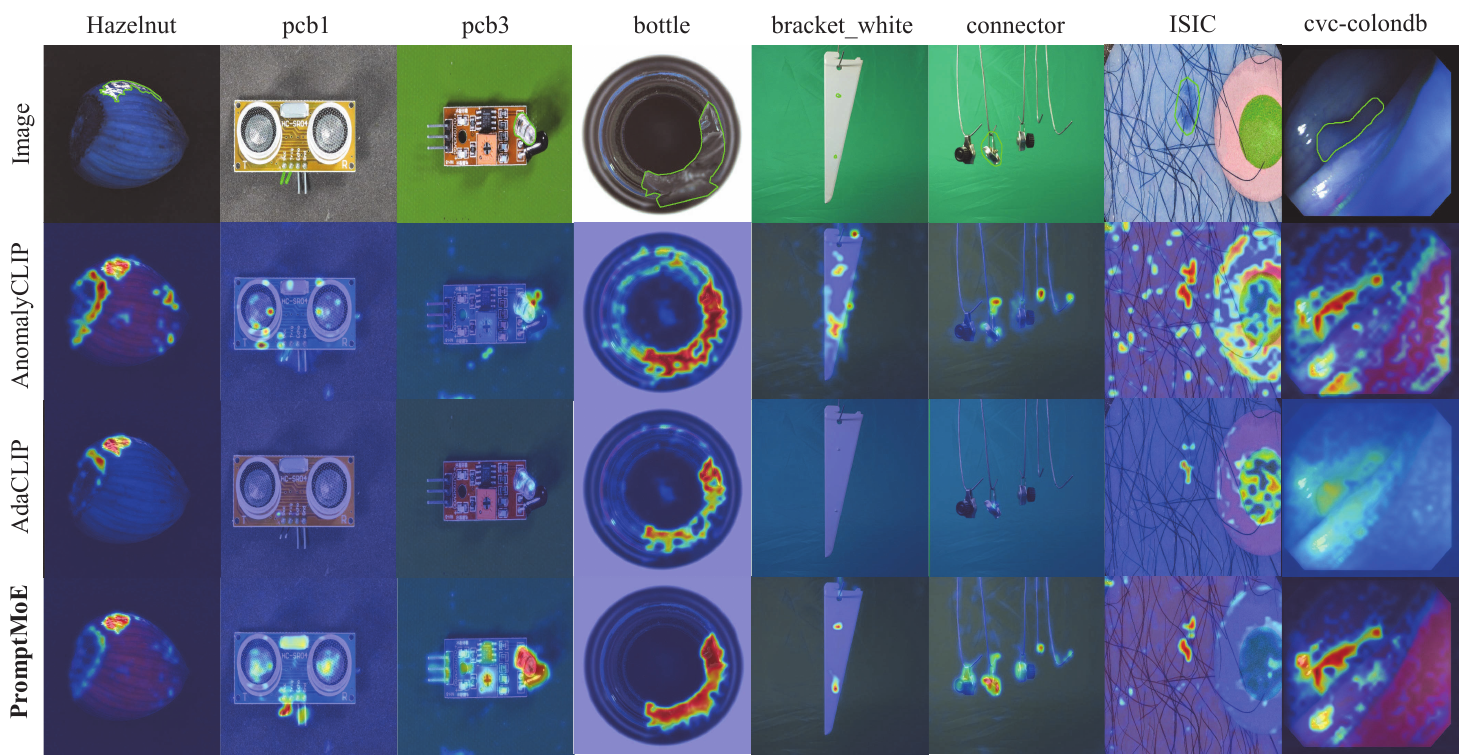}
	\caption{Qualitative comparison of anomaly localization results across different ZSAD methods.}
	\label{fig:visual}
\end{figure}

% \subsubsection{Comparing baselines.}

\subsection{Comparisons with State-of-the-Art Methods}
We compare $\mathtt{PromptMoE}$ against recent SOTA methods, including WinCLIP~\cite{winclip}, APRIL-GAN~\cite{chen2023zero}, AnomalyCLIP~\cite{anomalyclip}, AdaCLIP~\cite{adaclip}, and FAPrompt~\cite{zhu2024fine}. For a fair comparison, all results are cited from their original papers, while any missing values are reproduced using the official code under a unified setting.
\subsubsection{Quantitative Comparisons.}
 We quantitatively compare $\mathtt{PromptMoE}$ against recent SOTA methods across 15 industrial and medical datasets. The detailed results in Table~\ref{table:zsad_all}, visually summarized by the radar chart in Fig.~\ref{fig:radar}, clearly indicate that our $\mathtt{PromptMoE}$ demonstrates comprehensive superiority over competing methods in both image-level and pixel-level metrics. For instance, it achieves a 93.8\% Image-level AUROC on the MVTec AD benchmark (+1.8\% over the runner-up). This strong generalization is further validated on medical datasets, notably on HeadCT, where our model surpasses the next-best method by a significant 3.4\% in AUROC and 4.7\% in AP. The exceptional performance on the unseen medical domain, despite being trained solely on industrial data, strongly validates the success of our visually-guided compositional prompts in learning generalizable concepts of normality and abnormality.

\subsubsection{Qualitative Comparisons.}
To provide a visual intuition of the superiority of $\mathtt{PromptMoE}$, we present a series of anomaly localization results in Fig.~\ref{fig:visual}, comparing qualitatively against AnomalyCLIP and AdaCLIP. As can be observed, the results from competing methods are often imprecise, suffering from issues such as false positives on normal regions or incomplete coverage of defective areas. In contrast, the anomaly maps generated by our $\mathtt{PromptMoE}$ not only focus more accurately on the true defects but also exhibit highlighting that aligns more closely with the actual anomaly contours. These results suggest that our visually-guided compositional prompt mechanism, by generating more discriminative textual representations for each instance, enables a finer-grained image-text alignment that leads to superior localization performance.

\begin{figure}[htb]
        \centering
	\includegraphics[width=.90\linewidth]{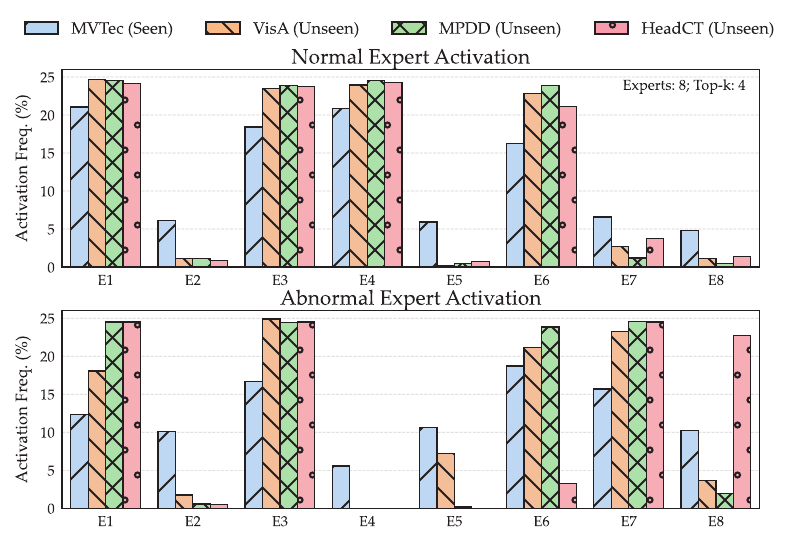}
	\caption{Activation frequencies of normal and abnormal experts across training and test datasets.}
	\label{fig:expert_activation}
\end{figure}

\subsubsection{Analysis of Expert Activation.}
To gain deeper insight into the dynamic routing behavior of the VGMoP module, we analyze its expert activation patterns. As illustrated in Fig.~\ref{fig:expert_activation}, we compute the cumulative activation frequency for each expert in both the normal and abnormal prompt pools on the seen dataset (MVTec AD) and across several unseen datasets.

The results clearly reveal that the normal and abnormal branches learn two distinct and highly task-adaptive routing strategies. For the normal state, the router's selections consistently converge to a small, fixed set of core experts across all test datasets. This indicates that our model successfully distills a core set of highly generalizable semantic primitives for the concept of ``normality'' from the diverse training data.

In stark contrast, the expert activation for the abnormal state remains highly dynamic and sparse across different unseen datasets, demonstrating that the model flexibly composes different expert primitives for varied anomaly patterns. For instance, we observe that a general-purpose expert (e.g., E3) is combined with a dataset-specific subset of other experts (e.g., higher activation of E7 and E8 on VisA vs. E1 and E6 on MPDD). Furthermore, the specific activation weights for the instances in Fig.~\ref{fig:teaser} reveal that the router assigns non-uniform weights to the selected experts, forming a meaningful weighted combination rather than a simple average. This provides further evidence of the router's nuanced, instance-specific decision-making.

% \subsubsection{Visualization of Compositional Representations.}

\subsection{Ablation Studies}
\subsubsection{Analysis of the Compositional Prompting.}
To validate the effectiveness of our proposed compositional prompting strategy, we conduct an ablation study progressively building from a simple static prompt baseline to our full $\mathtt{PromptMoE}$ model. As shown in Table~\ref{tab:components}, the ``Static Prompt'' baseline (using a single learnable prompt) yields limited performance. While ``Static Ensemble'', which uses a set of static prompts and averages their outputs, provides some improvement, its effectiveness remains constrained. Notably, replacing the static ensemble with ``VGMoP'' (single-layer, without auxiliary losses) leads to a significant performance boost (a +0.9\% gain in I-AUC on MVTec AD). This strongly demonstrates that the core advantage of our method stems from the dynamic, visually-guided composition of experts, rather than merely increasing the number of static prompts. Finally, our $\mathtt{PromptMoE}$, benefiting from its multi-layer prompt design and regularization from auxiliary losses, achieves the best performance, fully validating the superiority of our compositional framework.

\begin{table}[htbp]
	\small
	\centering
		\begin{tabular}{l  cc cc}
			\toprule
			\multirow{2}*{\textbf{Configuration}} & \multicolumn{2}{c}{MVTec AD} & \multicolumn{2}{c}{VisA}\\
			\cmidrule(lr){2-3} \cmidrule(lr){4-5}
			 & I-AUC & PRO & I-AUC & PRO \\
			\midrule
			Static Prompt& 91.7 & 82.0 & 82.4 & 88.0 \\
			\textit{+Static Ensemble}  & 92.2  & 82.9  & 83.3 & 88.3 \\
			 \textit{+VGMoP}  & \underline{93.1}  & \textbf{83.3}  & \underline{84.1} & \underline{89.0} \\
			% \midrule
			\midrule
			$\mathtt{PromptMoE}$  & \textbf{93.8}  & \underline{83.2}  & \textbf{85.0} & \textbf{89.2}\\
			\bottomrule
		\end{tabular}%

    \caption{Ablation on the effectiveness of compositional prompting strategy.}
	\label{tab:components}
\end{table}

% \subsubsection{Analysis of Expert Sharing Strategies.}
% Inspired by parameter sharing paradigms in multi-task learning, such as MMoE~\cite{ma2018modeling}, we further investigate the possibility of sharing components between the normal and abnormal state-prompting branches. Although our default design with separate components aims to ensure expert specialization through task isolation, a unified expert pool or a shared cross-attention mechanism could theoretically improve parameter efficiency by learning lower-level, shared semantic primitives. To validate the necessity and superiority of our separated design, we conduct the following ablation study.

% \begin{table}[htbp]
% 	\small
% 	\centering
% 	\caption{Ablation study on expert sharing strategies.}
% 		\begin{tabular}{l  cc cc}
% 			\toprule
% 			\multirow{2}*{\textbf{Configuration}} & \multicolumn{2}{c}{MVTec AD} & \multicolumn{2}{c}{VisA}\\
% 			\cmidrule(lr){2-3} \cmidrule(lr){4-5}
% 			 & I-AUC & PRO & I-AUC & PRO \\
% 			\midrule
% 			\textit{Separate Components}  & 93.8  & 83.2  & 85.0 & 89.2\\
%                 \midrule
% 			\textit{w/ Shared Prompt Pool}  & 98.8  & 97.4  & 93.3 & \\
% 			 \textit{w/ Shared Cross-Atte.}  & 99.0  & 97.9  & 94.4 & \\
% 			 \textit{w/ Shared Both}  & 99.0  & 97.9  & 94.4 & \\

% 			\bottomrule
% 		\end{tabular}%
% 	\label{tab:share}
% \end{table}

\begin{figure}[htb]
        \centering
	\includegraphics[width=.85\linewidth]{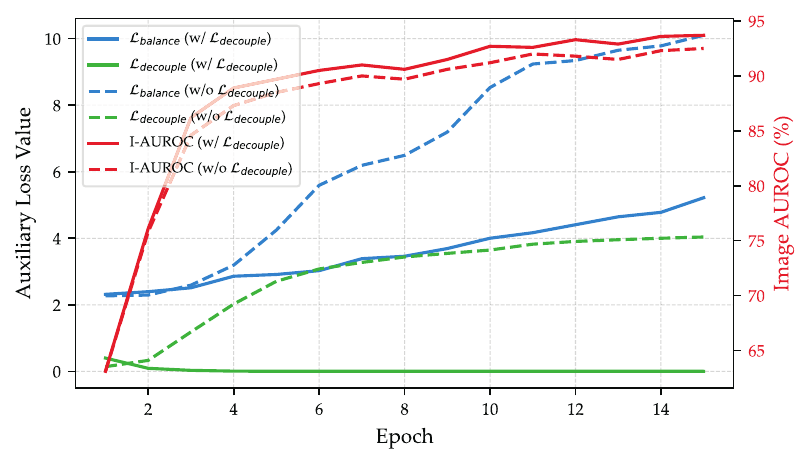}
	\caption{Effect of $\mathcal{L}_{decouple}$ loss on MoE training dynamics.}
	\label{fig:aux_loss_ablation}
\end{figure}

\subsubsection{Synergy of Expert Diversity and Load Balancing.}
To validate the synergy between our two auxiliary losses, we visualize the evolution of key metrics during training with and without $\mathcal{L}_\text{decouple}$ in Fig.~\ref{fig:aux_loss_ablation}. The results clearly show that in the absence of $\mathcal{L}_\text{decouple}$ (dashed lines), expert representations become redundant (reflected by the increasing value of $\mathcal{L}_\text{decouple}$), which in turn leads to a sharp degradation of $\mathcal{L}_\text{balance}$ and ultimately harms the model's I-AUC performance. Conversely, when both losses are used (solid lines), both auxiliary losses are effectively suppressed to healthy levels, and the model achieves superior final performance. This strongly demonstrates that the expert diversity enforced by $\mathcal{L}_\text{decouple}$ is a crucial prerequisite for achieving effective load balancing and optimal performance.

\subsubsection{Analysis on the Load Balancing loss.}
To further validate the necessity of $\mathcal{L}_\text{balance}$, we conduct an ablation study on its coefficient $\alpha$. As shown in Table~\ref{tab:ablation_balance}, completely removing the load balancing loss ($\alpha=0$) leads to a significant performance drop. A moderate weight ($\alpha=0.01$) effectively promotes expert balancing and yields the best performance. However, an excessively high weight ($\alpha=0.1$) might over-penalize the router's specialization, which also negatively impacts the main task's performance.

\begin{table}[htbp]
	\small
	\centering
	\begin{tabular}{l cc cc}
		\toprule
		\multirow{2}{*}{\textbf{Weight $\alpha$}} & \multicolumn{2}{c}{MVTec AD} & \multicolumn{2}{c}{VisA}\\
		\cmidrule(lr){2-3} \cmidrule(lr){4-5}
		 & I-AUC & PRO & I-AUC & PRO \\
		\midrule
		$\alpha = 0$ (w/o $\mathcal{L}_\text{balance}$) & 92.1 & 82.5 & 83.4 & 88.2 \\
		$\alpha = 0.01$ & \textbf{93.8} & \textbf{83.2} & \textbf{85.0} & \textbf{89.2} \\
		$\alpha = 0.1$ & 92.9 & 83.1 & 84.2 & \textbf{89.2} \\
		\bottomrule
	\end{tabular}
	\caption{Ablation study on the weight $\alpha$ of $\mathcal{L}_\text{balance}$.}
    \label{tab:ablation_balance}

\end{table}

\section{Conclusion}
In this paper, we have demonstrated that reframing prompt learning as a compositional paradigm is an effective approach to tackling the generalization challenge in ZSAD. Our proposed $\mathtt{PromptMoE}$, through its visually-guided MoE mechanism, successfully learns to dynamically construct tailored textual prompts for each instance from a basis set of learnable semantic primitives. This approach overcomes the representational bottlenecks and overfitting risks associated with monolithic prompts. The SOTA performance achieved across 15 industrial and medical datasets strongly supports our central conclusion: learning how to compose, rather than learning a single complete prompt, is key to enhancing a model's ability to handle the diversity of unseen anomalies.

\section*{Acknowledgments}
This work was supported in part by the National Key Research and
Development Program of China (2022YFB4501704); the
National Science Foundation of China (62222312, 62473285); the Shanghai Science and Technology Innovation Action Plan
Project of China (22511100700); and the Fundamental Research
Funds for the Central Universities.

\bibliography{aaai2026}

\clearpage

\appendix
\section*{Appendix}
This supplementary material provides additional details and analyses for our work. Specifically, we supplement with: (A) complete implementation details; (B) additional ablation studies to validate the effectiveness of our key design choices; (C) visualization analysis of expert semantics; (D) detailed statistics for all datasets used; and (E) more exhaustive, category-level performance results on key benchmarks.

\section{A. Implementation details}
We utilize the pre-trained CLIP model (ViT-L/14@336px) released by OpenAI~\cite{radford2021learning} as our backbone, with all of its parameters frozen during training. All input images are resized to $518 \times 518$ pixels. We extract patch features from the \{6, 12, 18, 24\}-th layers of the 24-layer visual encoder. The model is trained for 15 epochs using the Adam optimizer with betas set to $(0.6, 0.999)$ and a batch size of 16. The learning rate is scheduled with a linear warm-up for the first 3 epochs, reaching a final value of 0.001. For our VGMoP module, we set the number of queries $N_q=8$ and experts $E=8$ for both the normal and abnormal branches, from which we sparsely select the top $k=4$ for aggregation; its cross-attention utilizes 8 heads and its router has a hidden dimension of 256. The resulting normal and abnormal state sequences have lengths of $M_n=5$ and $M_a=6$, respectively, while the learnable shared context $\mathbf{Q}_{ctx}$ has a length of $M_q=8$. The coefficients for the auxiliary losses, $\alpha$ for $\mathcal{L}_\text{balance}$ and $\beta$ for $\mathcal{L}_\text{decouple}$, are set to 0.01 and 0.005, respectively. For similarity calculations, the temperature parameters for the pixel-level $\tau$ and image-level $\tau'$ computations are set to 0.07 and 0.01. During inference, the final pixel-level anomaly map $\mathbf{M}$ is smoothed with a Gaussian filter with a sigma of 4. All experiments are conducted on a NVIDIA GeForce RTX 3090 GPU using PyTorch 1.13.1.

\section{B. Additional Ablations}
\subsubsection{Analysis on Visual Feature Layers}
We investigate the impact of using different combinations of feature layers from the visual encoder, with results presented in Table~\ref{tab:ablation_layers}. The experiment clearly indicates that fusing multi-level features consistently improves overall performance compared to using only the final layer's (L24) features. The strategy of utilizing all four layers (\{6, 12, 18, 24\}) achieves the most robust and best overall performance across both datasets. 

\begin{table}[htbp]
	\small
	\centering
	\caption{Ablation study on the set of visual feature layers. We report average Image-level AUROC and Pixel-level PRO score (\%).}
		\begin{tabular}{cccc  cc cc}
			\toprule
			\multicolumn{4}{c}{\textbf{Selected Layers}} & \multicolumn{2}{c}{MVTec AD} & \multicolumn{2}{c}{VisA}\\
			\cmidrule(lr){1-4} \cmidrule(lr){5-6} \cmidrule(lr){7-8}
			 6 & 12 & 18 & 24 & I-AUC & PRO & I-AUC & PRO \\
			\midrule
			& & & \Checkmark  & 93.2  & 83.2  & 83.9 & 89.1 \\
                % \Checkmark & & & \Checkmark  & 93.5  & 84.9  & & \\
			 & \Checkmark & & \Checkmark  & 93.3 & 84.1& 84.0 & 89.1 \\
			% & & \Checkmark & \Checkmark  & 93.5  & 84.9  &  & \\
			\Checkmark & \Checkmark & & \Checkmark  &  93.4 & 83.1 & 84.7 & 89.3\\
			\Checkmark & & \Checkmark & \Checkmark  &  93.5 & 83.2 & 84.6 & 89.2\\
			\Checkmark & \Checkmark & \Checkmark & \Checkmark  & 93.8  & 83.2  & 85.0 & 89.2 \\

			\bottomrule
		\end{tabular}%
	\label{tab:ablation_layers}
\end{table}

\subsubsection{Analysis of Expert Sharing Strategies.}
Inspired by parameter sharing paradigms in multi-task learning, such as MMoE~\cite{ma2018modeling}, we further investigate the possibility of sharing components between the normal and abnormal state-prompting branches. Although our default design with separate components aims to ensure expert specialization through task isolation, a unified expert pool or a shared cross-attention mechanism could theoretically improve parameter efficiency by learning lower-level, shared semantic primitives. To validate the necessity and superiority of our separated design, we conduct the following ablation study.

\begin{table}[thbp]
	\small
	\centering
	\caption{Ablation study on expert sharing strategies.}
		\begin{tabular}{l  cc cc}
			\toprule
			\multirow{2}*{\textbf{Configuration}} & \multicolumn{2}{c}{MVTec AD} & \multicolumn{2}{c}{VisA}\\
			\cmidrule(lr){2-3} \cmidrule(lr){4-5}
			 & I-AUC & PRO & I-AUC & PRO \\
			\midrule
			\textit{Separate Components}  & 93.8  & 83.2  & 85.0 & 89.2\\
                \midrule
			\textit{w/ Shared Prompt Pool}  & 91.4  & 82.8  & 83.3 & 87.8\\
			 \textit{w/ Shared Cross-Atte.}  & 92.9  & 83.1  & 84.2 & 88.8\\
			 \textit{w/ Shared Both}  & 90.9  & 82.5  & 82.6 & 87.9\\

			\bottomrule
		\end{tabular}%
	\label{tab:share}
\end{table}

As shown in Table~\ref{tab:share}, the results clearly indicate that all forms of parameter sharing lead to a performance drop compared to our default Separate Components design. Specifically, sharing either the cross-attention module (`w/ Shared Cross-Atte.') or the expert prompt pool (`w/ Shared Prompt Pool') impairs performance, while sharing both components (`w/ Shared Both') results in the most significant degradation. This outcome strongly validates our core design hypothesis: providing independent and specialized representational pathways for the semantically adversarial tasks of modeling normality and abnormality is crucial for avoiding negative transfer and achieving optimal zero-shot detection performance.
\subsubsection{Analysis of MoE Hyperparameters.}
We further investigate the impact of two key hyperparameters in our VGMoP module: the total number of experts ($E$) and the number of selected experts ($k$). As shown in Table~\ref{tab:hyperparams_moe}, we observe that increasing the total number of experts (from 8 to 16) provides a richer pool of semantic primitives, which leads to consistent performance gains in pixel-level metrics PRO. However, this does not always translate to a corresponding improvement in image-level metrics I-AUC. Furthermore, the results indicate that selecting half of the experts ($k/E = 50\%$) achieves the best performance balance. Considering all factors, our final model adopts the $E=8, k=4$ configuration, as it provides the most robust and efficient performance across all metrics.

\begin{table}[bthp]
	\centering
        \small
	\caption{Ablation study on the number of experts ($E$) and selected experts ($k$) for VGMoP.}
	\label{tab:hyperparams_moe}
	\begin{tabular}{cc cc cc}
		\toprule
		\multicolumn{2}{c}{\textbf{Parameters}} & \multicolumn{2}{c}{MVTec AD} & \multicolumn{2}{c}{VisA}\\
		\cmidrule(r){1-2} \cmidrule(r){3-4} \cmidrule(r){5-6}
		$E$ & $k$ & I-AUC & PRO & I-AUC & PRO \\
        \midrule
        4 & 4 & 92.5 & 83.0 & 83.7 & 88.6 \\
		\midrule
		\multirow{3}{*}{8} 
		 & 1 & 92.0 & 83.0 & 84.0 & 89.0 \\
		 & 2 & 92.6 & 83.2 & 84.3 & 89.0 \\
		 & 4 & \textbf{93.8} & 83.2 & \textbf{85.0} & 89.2 \\
		\midrule
		\multirow{3}{*}{16} 
		 & 2 & 92.2 & 82.8 & 84.1 & 88.9 \\
		 & 4 & 92.9 & 82.9 & 84.2 & 89.1 \\
          & 8 & 93.0 & \textbf{84.0} & 84.7 & \textbf{89.3} \\

		\bottomrule
	\end{tabular}
\end{table}

\section{C. Visualization}
\subsubsection{Analysis of Expert Semantics.}
\begin{figure}[tbhp]
        \centering
	\includegraphics[width=1.\linewidth]{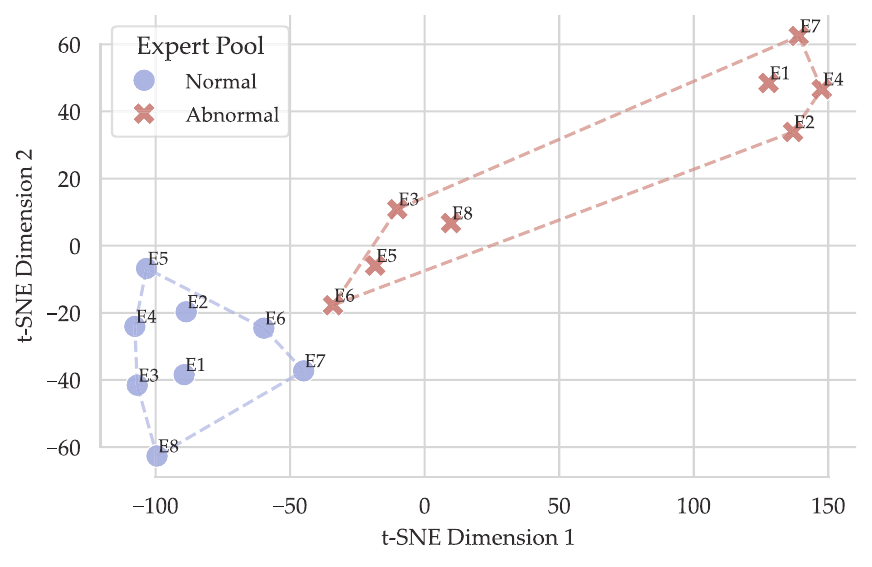} % 请确保路径正确
	\caption{t-SNE visualization of the mean embeddings of experts from the Normal and Abnormal prompt pools. Each point represents a single expert.}
	\label{fig:expert_tsne_visualization}
\end{figure}

To gain deeper insight into the learned semantic structure of our expert prompt pools, we visualize the mean embeddings of each expert from both the normal and abnormal pools using t-SNE, as shown in Fig.~\ref{fig:expert_tsne_visualization}.

The results clearly reveal two distinct semantic distributions for the expert pools. The normal experts form a compact cluster in the feature space, which indicates that our model has learned a semantically consistent set of core representations for the relatively convergent concept of ``normality''. In stark contrast, the abnormal experts are highly dispersed, occupying a broad area of the feature space. This strongly demonstrates that our method has successfully learned a diverse and discriminative set of ``semantic primitives'' for the open-ended and varied concept of ``abnormality''. This observed pattern of ``normal converging, abnormal diverging'' not only visually validates the effectiveness of our independent expert pool design but also showcases the significant potential of our compositional framework for handling unseen anomalies.

\section{D. Datasets}
To comprehensively evaluate the generalization capability of our proposed $\mathtt{PromptMoE}$ on the ZSAD task, we utilize 15 public datasets spanning industrial manufacturing and medical diagnosis domains. As detailed in Table~\ref{table:dataset_details}, these datasets encompass diverse object and texture classes, different imaging modalities, and a wide range of anomaly patterns. 
\begin{table*}[tbhp]
    \small
    \centering
    \caption{Details of the 15 datasets used in our experiments. For ZSAD, we only utilize the test splits of these datasets. Sample counts for MVTec AD and VisA reflect their test sets.}
    \label{table:dataset_details}
    \resizebox{\textwidth}{!}{%
    \begin{tabular}{cllccc}
        \toprule
        \textbf{Domain} & \textbf{Dataset} & \textbf{Type} & \textbf{Modalities} & \textbf{Categories ($|\mathcal{C}|$)} & \textbf{Samples (Normal, Anomalous)} \\
        \midrule
        \multirow{7}{*}{\rotatebox[origin=c]{90}{Industrial}} 
        & MVTec AD~\cite{bergmann2019mvtec} & Object \& Texture & Photography & 15 & (467, 1258) \\
        & VisA~\cite{zou2022spot} & Object & Photography & 12 & (962, 1200) \\
        & MPDD~\cite{jezek2021deep} & Object & Photography & 6 & (176, 282) \\
        & BTAD~\cite{mishra2021vt} & Object & Photography & 3 & (451, 290) \\
        & SDD~\cite{tabernik2020segmentation} & Object & Photography & 1 & (181, 74) \\
        & DAGM~\cite{wieler2007weakly} & Texture & Photography & 10 & (6996, 1054) \\
        & DTD-Synthetic~\cite{aota2023zero} & Texture & Photography & 12 & (357, 947) \\
        \midrule
        \multirow{8}{*}{\rotatebox[origin=c]{90}{Medical}}
        & HeadCT~\cite{salehi2021multiresolution} & Brain & Radiology (CT) & 1 & (100, 100) \\
        & BrainMRI~\cite{kanade2015brain} & Brain & Radiology (MRI) & 1 & (98, 155) \\
        & Br35H~\cite{hamada2020br35h} & Brain & Radiology (MRI) & 1 & (1500, 1500) \\
        & ISIC~\cite{codella2018skin} & Skin & Photography & 1 & (0, 379) \\
        & CVC-ColonDB~\cite{tajbakhsh2015automated} & Colon & Endoscopy & 1 & (0, 380) \\
        & CVC-ClinicDB~\cite{bernal2015wm} & Colon & Endoscopy & 1 & (0, 612) \\
        & Kvasir~\cite{jha2019kvasir} & Colon & Endoscopy & 1 & (0, 1000) \\
        & Endo~\cite{hicks2021endotect} & Colon & Endoscopy & 1 & (0, 200) \\
        \bottomrule
    \end{tabular}%
    }
\end{table*}

\section{E. Detailed Results}

For a more detailed comparison, this section presents category-level results of our $\mathtt{PromptMoE}$ on key benchmarks including MVTec AD~\cite{bergmann2019mvtec}, VisA~\cite{zou2022spot}, MPDD~\cite{jezek2021deep}, BTAD~\cite{mishra2021vt}, and DTD-Synthetic~\cite{aota2023zero} in Table~\ref{table:mvtec_detailed}-\ref{table:dtd_detailed}.

\begin{table*}[thbp]
    \centering
    \caption{Category-level performance on the MVTec AD dataset. We report Image-level (I-AUROC, I-AP) and Pixel-level (P-AUROC, PRO) scores (\%).}
    \label{table:mvtec_detailed}
    \begin{tabular}{l cc cc}
        \toprule
        \multirow{2}{*}{\textbf{Category}} & \multicolumn{2}{c}{\textbf{Image-level}} & \multicolumn{2}{c}{\textbf{Pixel-level}} \\
        \cmidrule(lr){2-3} \cmidrule(lr){4-5}
         & I-AUROC & I-AP & P-AUROC & PRO \\
        \midrule
        \multicolumn{5}{l}{\textit{Textures}} \\
        \midrule
        Carpet     & 100 & 100 & 99.0 & 90.6 \\
        Grid       & 99.6  & 99.8  & 97.9 & 75.9 \\
        Leather    & 100.0 & 100.0 & 99.2 & 96.9 \\
        Tile       & 99.2  & 99.7  & 96.4 & 73.0 \\
        Wood       & 97.8  & 99.4  & 97.4 & 94.5 \\
        \midrule
        \multicolumn{5}{l}{\textit{Objects}} \\
        \midrule
        Bottle     & 93.8  & 98.2  & 91.0 & 83.8 \\
        Cable      & 82.6  & 89.5  & 80.0 & 70.0 \\
        Capsule    & 91.3  & 98.2  & 96.2 & 87.0 \\
        Hazelnut   & 96.9  & 98.3  & 96.9 & 95.4 \\
        Metal Nut  & 95.7  & 99.0  & 79.1 & 80.6 \\
        Pill       & 84.1  & 96.8  & 87.6 & 92.4 \\
        Screw      & 87.1  & 94.4  & 98.3 & 84.9 \\
        Toothbrush & 89.4  & 95.0  & 93.5 & 90.7 \\
        Transistor & 92.5  & 90.7  & 68.8 & 55.1 \\
        Zipper     & 96.6  & 99.0  & 95.5 & 77.2 \\
        \midrule
        \textbf{Average} & 93.8 & 97.2 & 91.8 & 83.2 \\
        \bottomrule
    \end{tabular}
\end{table*}

\begin{table*}[thbp]
    \centering
    \caption{Category-level performance on the VisA dataset.}
    \label{table:visa_detailed}
    \begin{tabular}{l cc cc}
        \toprule
        \multirow{2}{*}{\textbf{Category}} & \multicolumn{2}{c}{\textbf{Image-level}} & \multicolumn{2}{c}{\textbf{Pixel-level}} \\
        \cmidrule(lr){2-3} \cmidrule(lr){4-5}
         & I-AUROC & I-AP & P-AUROC & PRO \\
        \midrule
        Candle     & 87.4 & 90.2 & 98.8 & 95.1 \\
        Capsules   & 72.2 & 84.0 & 96.8 & 87.5 \\
        Cashew     & 82.3 & 92.5 & 88.3 & 93.2 \\
        Chewinggum & 97.6 & 99.0 & 99.1 & 91.5 \\
        Fryum      & 92.1 & 96.2 & 96.1 & 88.4 \\
        Macaroni1  & 88.9 & 89.5 & 99.2 & 92.9 \\
        Macaroni2  & 74.6 & 74.7 & 98.7 & 89.6 \\
        PCB1       & 91.0 & 91.2 & 94.0 & 87.8 \\
        PCB2       & 75.6 & 75.4 & 93.4 & 82.2 \\
        PCB3       & 66.3 & 71.5 & 89.3 & 76.2 \\
        PCB4       & 95.8 & 95.2 & 94.2 & 90.4 \\
        Pipe Fryum & 96.4 & 98.3 & 98.8 & 95.5 \\
        \midrule
        \textbf{Average} & 85.0 & 88.1 & 95.6 & 89.2 \\
        \bottomrule
    \end{tabular}
\end{table*}

\begin{table*}[thbp]
    \centering
    \caption{Category-level performance on the MPDD dataset.}
    \label{table:mpdd_detailed}
    \begin{tabular}{l cc cc}
        \toprule
        \multirow{2}{*}{\textbf{Category}} & \multicolumn{2}{c}{\textbf{Image-level}} & \multicolumn{2}{c}{\textbf{Pixel-level}} \\
        \cmidrule(lr){2-3} \cmidrule(lr){4-5}
         & I-AUROC & I-AP & P-AUROC & PRO \\
        \midrule
        Bracket black & 80.0 & 81.4 & 97.8 & 92.6 \\
        Bracket brown & 63.3 & 76.0 & 94.6 & 87.1 \\
        Bracket white & 91.1 & 92.6 & 99.8 & 96.2 \\
        Connector     & 74.8 & 63.0 & 95.2 & 81.1 \\
        Metal plate   & 88.7 & 95.9 & 94.7 & 81.2 \\
        Tubes         & 95.8 & 98.3 & 98.6 & 94.9 \\
        \midrule
        \textbf{Average} & 82.3 & 84.5 & 96.8 & 88.8 \\
        \bottomrule
    \end{tabular}
\end{table*}

\begin{table*}[thbp]
    \centering
    \caption{Category-level performance on the BTAD dataset.}
    \label{table:btad_detailed}
    \begin{tabular}{l cc cc}
        \toprule
        \multirow{2}{*}{\textbf{Category}} & \multicolumn{2}{c}{\textbf{Image-level}} & \multicolumn{2}{c}{\textbf{Pixel-level}} \\
        \cmidrule(lr){2-3} \cmidrule(lr){4-5}
         & I-AUROC & I-AP & P-AUROC & PRO \\
        \midrule
        01 & 98.8 & 99.5 & 95.4 & 77.8 \\
        02 & 83.4 & 97.4 & 93.2 & 72.4 \\
        03 & 98.1 & 90.3 & 96.1 & 91.6 \\
        \midrule
        \textbf{Average} & 93.4 & 95.7 & 94.9 & 80.6 \\
        \bottomrule
    \end{tabular}
\end{table*}

\begin{table*}[thbp]
    \centering
    \caption{Category-level performance on the DTD-Synthetic dataset.}
    \label{table:dtd_detailed}
    \begin{tabular}{l cc cc}
        \toprule
        \multirow{2}{*}{\textbf{Category}} & \multicolumn{2}{c}{\textbf{Image-level}} & \multicolumn{2}{c}{\textbf{Pixel-level}} \\
        \cmidrule(lr){2-3} \cmidrule(lr){4-5}
         & I-AUROC & I-AP & P-AUROC & PRO \\
        \midrule
        Woven 001       & 100.0 & 100.0 & 99.8 & 99.2 \\
        Woven 127       & 87.9  & 89.6  & 94.1 & 89.4 \\
        Woven 104       & 98.6  & 99.7  & 95.8 & 92.0 \\
        Stratified 154  & 100.0 & 100.0 & 99.4 & 99.2 \\
        Blotchy 099     & 100.0 & 100.0 & 99.6 & 96.8 \\
        Woven 068       & 96.8  & 98.1  & 99.0 & 92.1 \\
        Woven 125       & 100.0 & 100.0 & 99.7 & 96.8 \\
        Marbled 078     & 99.2  & 99.8  & 99.3 & 97.7 \\
        Perforated 037  & 94.6  & 98.8  & 96.2 & 89.7 \\
        Mesh 114        & 87.2  & 94.7  & 95.5 & 79.3 \\
        Fibrous 183     & 98.5  & 99.7  & 99.4 & 98.6 \\
        Matted 069      & 88.0  & 96.8  & 99.5 & 87.6 \\
        \midrule
        \textbf{Average} & 95.9 & 98.1 & 98.3 & 93.2 \\
        \bottomrule
    \end{tabular}
\end{table*}

\end{document}